# Machine learning techniques to identify antibiotic resistance in patients diagnosed with various skin and soft tissue infections

Farnaz H. Foomani[1], Shahzad Mirza[2], Sahjid Mukhida[2], Kannuri Sriram[2], Zeyun Yu[1], Aayush Gupta[2], and Sandeep Gopalakrishnan[3]

[1]Department of Electrical Engineering and Computer Science, University of Wisconsin-Milwaukee, Milwaukee, WI, United States, [2]Department of Microbiology and Department of Dermatology, Dr. D.Y. Patil Medical College Hospital and Research Centre, Dr D Y Patil Vidyapeeth, Pimpri, Pune, India. [3]College of Nursing, University of Wisconsin-Milwaukee, Milwaukee, WI, United States.

## Abstract:

Skin and soft tissue infections (SSTIs) are among the most frequently observed diseases in ambulatory and hospital settings. Resistance of diverse bacterial pathogens to antibiotics is a significant cause of severe SSTIs, and treatment failure results in morbidity, mortality, and increased cost of hospitalization. Therefore, antimicrobial surveillance is essential to predict antibiotic resistance trends and monitor the results of medical interventions. To address this, we developed machine learning (ML) models (deep and conventional algorithms) to predict antimicrobial resistance using antibiotic susceptibility testing (ABST) data collected from patients clinically diagnosed with primary and secondary pyoderma over a period of one year. We trained an individual ML algorithm on each antimicrobial family to determine whether a Gram-Positive Cocci (GPC) or Gram-Negative Bacilli (GNB) bacteria will resist the corresponding antibiotic. For this purpose, clinical and demographic features from the patient and data from ABST were employed in training. We achieved an Area Under the Curve (AUC) of 0.68-0.98 in GPC and 0.56-0.93 in GNB bacteria, depending on the antimicrobial family. We also conducted a correlation analysis to determine the linear relationship between each feature and antimicrobial families in different bacteria. ML techniques suggest that a predictable nonlinear relationship exists between patients' clinical-demographic characteristics and antibiotic resistance; however, the accuracy of this prediction depends on the type of the antimicrobial family.

**Co- Corresponding Authors:**

Shahzad Mirza, MD, Department of Microbiology, Dr, D.Y. Patil Medical College, Hospital and Research Center, Dr D Y Patil Vidyapeeth, Pimpri, Pune 411018, India

Sandeep Gopalakrishnan, Ph.D., DAPWCA, College of Nursing, University of Wisconsin-Milwaukee, Milwaukee, WI, United States

## Introduction

Skin and soft-tissue infections (SSTIs) remain the most frequently observed infections in ambulatory and hospital settings. In 2005 alone, approximately 14.2 million SSTI cases were reported in the United States, a 65% increase from 1997[1]. The bacteria causing SSTIs are commonly classified into Gram-negative or Gram-positive depending on their strain characteristic. Multi-drug resistant Gram-negative bacteria (GNB), especially *Enterobacteriaceae* and the non-fermenters, cause significant public health problems due to their challenging treatment resulting in increased morbidity and mortality along with a prolonged intensive care unit (ICU) stay [2]. Similarly, Gram-positive organisms like, *Staphylococcus aureus, Streptococcus species, Enterococcus species, etc.*, may also exhibit a variety of resistance and virulence factors that contribute to their prominent role in causing infections of the critically ill, especially in the hospital setting[3]. *Staphylococci* are transmitted from human to human; the organisms may form a part of the patient's normal commensal flora and be introduced later to any sterile site by invasive procedures or traumatic injury. The human transmission of resistant strains like Methicillin-Resistant Staphylococcus aureus (MRSA) commonly occurs in a hospital environment but may rarely be encountered in the community setting. The SCOPE project (Surveillance and Control of Pathogens of Epidemiologic Importance) found that GPC accounted for 62 percent of all bloodstream infections in 1995 and 76 percent in 2000, while GNB accounted for 22 percent in 1995 and 14 percent in 2000,in individuals with an underlying malignancy [4].

The most common community acquired SSTI include cellulitis, folliculitis, furunculosis, and secondary infected traumatic ulcers. The recent increase in the number of invasive medical procedures and immunocompromised patients due to immunosuppressive drugs, cancer, transplant surgery, and HIV/AIDS have dramatically increased the incidence of serious SSTI[4]. Although most cases are mild and can be treated with empirical oral antimicrobial agents, especially in India, moderate or severe circumstances of SSTI require hospitalization and parenteral therapy [5].

Since the introduction of antibiotics, an alarming increase in the resistance of diverse bacterial pathogens, including community-acquired organisms, has been observed [6]. The prevalence of multidrug resistance (MDR) has increased significantly among many pathogens due to the overuse and misuse of antimicrobial agents[7]. Such (mis)use increases the cost of medical care, exposes the patient to potential adverse effects, and risks the development and spread of antimicrobial resistance in healthcare facilities, significantly affecting our ability to treat patients empirically.

Hence, surveillance of antimicrobial resistance is essential for estimating the magnitude and trends of antibiotic resistance and monitoring the results of medical interventions[8]. Local surveillance data are critical and need to be used to direct clinical management, formulate treatment guidelines, and guide infection control policies[9]. However, there is a relative dearth of local antibiotic susceptibility data in patients with SSTI from India. Therefore, this study was carried out in a tertiary health care center in Western Maharashtra to assess the magnitude and clinical pattern of pyodermas', their causative microorganisms, and the antibiotic susceptibility patterns by using machine learning (ML).

ML techniques have shown their diagnostic and prognostic power in healthcare. Conventional ML algorithms such as decision trees and random forests are widely used in medicine due to their ability to analyze the relative importance of the input variables. However, these models are highly dependent on feature representations and are incapable of simulating the complexity of decision-making of a human neuronal system. Deep models, on the other hand, are inspired by the multi-level cognition of the human brain and have proved to be able to model the nonlinearity and complexity of human thinking. Deep neural network models have the potential of automatic feature extraction and can abstract high-level representations from low-level information.

Predicting antimicrobial resistance before culture using ML techniques can aid clinicians by increasing the effectiveness of empirical therapies. Moreover, they can help patients minimize their healthcare costs associated with culture and prolonged hospital stay. Antimicrobial resistance research based on ML algorithms has been developed for patients with urinary infection and ICU patients. In this article, we developed machine learning (ML) models (both deep and conventional algorithms) to predict antimicrobial resistance using antibiotic susceptibility testing (ABST) data collected from patients clinically diagnosed with SSTIs over one year. The networks learn from patients' clinical and demographic information such as age, gender, diagnoses, and bacterial pathogens involved with the skin infections. We also studied the linear relationship between each feature and antimicrobial families in GPC and GNB bacteria using correlation analysis and random forest.

## Methodology

### Dataset

We analyzed clinically relevant data of patients clinically diagnosed with primary and secondary pyoderma collected from the Departments of Dermatology and Microbiology of a tertiary care centre in Pune, India over one year. The dataset of 103 patients with GPC bacteria contains the variables of age in years (numerical), gender (binary), MRSA screening test (ordinal), Inducible clindamycin resistance (ordinal), Organism (categorical), and diagnosis (categorical). The class attribute, antibiotic resistance, is binary for Gentamicin, Cotrimoxazole, Cefoxitin, Erythromycin, Clindamycin, and Ciprofloxacin antibiotics. Table 1includes the summary statistics of the GPC dataset, and Figure 1represents a bar graph showing the distribution of each class in six antibiotic families. The dataset of 107 patients with GNB consists of age in years (numerical), gender (binary), ESBL test (ordinal), Carbapenems (ordinal), Organism (categorical), and diagnosis (categorical). The class labels indicate antibiotic resistance are binary for Gentamicin, Ceftazidime, Ceftazidime-Clavulanic Acid, Imipenem, Piperacillin-Tazobactam, Colistin, Amikacin, Ofloxacin, and Meropenem antibiotics. Table 2 shows the summary statistics of the GNB dataset, and Figure 2 includes the bar graphs of nine antibiotics susceptible/resistance distributions.

### Correlation Analysis

Correlation analysis is widely utilized to find interesting relationships in data. These relationships assist us in understanding the underlying relevance of attributes contributing to the prediction of the target class [10, 11]. To identify the relevant characteristics in the dataset which have a significant impact on the classification of antibiotic resistance, we used Pearson correlation analysis in numerical, Spearman correlation analysis in ordinal, and Cramer's V analysis in binary and categorical variables.

Table 1- Summary statistics of the GPC dataset.

| GPC (Gram Positive Cocci Bacteria) | | | | |
|---|---|---|---|---|
| Feature | Type | Feature | Type | |
| **Age (Years)** | Mean: 44.34<br>Std: 15.74<br>Range: 95 | **Organism** | ▪ *Staphylococcus aureus* (82.52%)<br>▪ *Enterococcus spp* (1.94%)<br>▪ *Streptococcus pyogenes* (5.8%)<br>▪ *Staphylococcus, Coagulase-negative* (9.7%) | |
| **Sex** | Male (65%)<br>Female (35%) | **Diagnosis** | ▪ Psoriasis (0.97%)<br>▪ Erythema (0.97%)<br>▪ Erythrasma (1.94%)<br>▪ Folliculitis (4.85%)<br>▪ Furuncle (1.94%)<br>▪ Hansen (15.53%)<br>▪ Infected Ulcer (0.97%)<br>▪ Impetigo (6.79%)<br>▪ Lichen (0.97%)<br>▪ Lupus (0.97%)<br>▪ Cellulitis (0.97%)<br>▪ Stasis ulcer | ▪ Mycetoma (1.94%)<br>▪ Pemphigus (6.79%)<br>▪ Pyoderma (5.82%) Gangrenosum<br>▪ Pyoderma (18.44%)<br>▪ Secondary infected eczema (4.85%)<br>▪ Sclerosis (0.97%)<br>▪ Toxic Necrolysis (0.97%)<br>▪ Abscess (3.88%)<br>▪ Burn (2.91%) |
| **MRSA screening test** | ▪ Positive (38.83%)<br>▪ Negative (43.68%)<br>▪ Not applicable (17.46%) | | | |
| **Inducible clindamycin resistance** | ▪ Positive (25.24%)<br>▪ Negative (74.76%) | | | |
| | | | (0.97%)<br>▪ Trophic ulcer (0.97%)<br>▪ Traumatic ulcer (0.97%) | ▪ Ecthyma (1.94%)<br>▪ Sebaceous cyst (1.94%)<br>▪ Vascular ulcer (0.97%)<br>▪ Eczema (8.73%) |

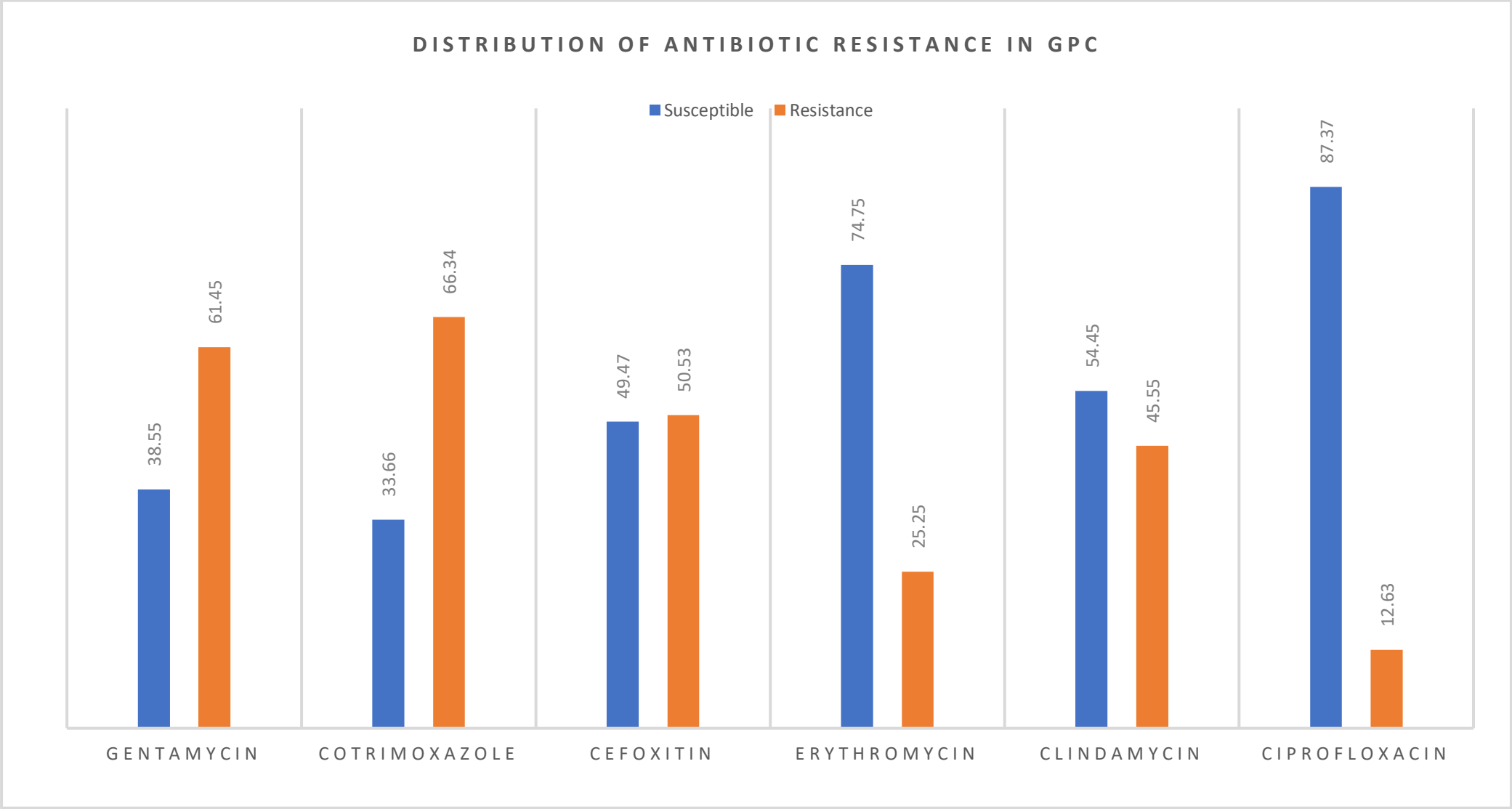

Figure 1- Percentage of GPC bacteria susceptible or resistant to six antimicrobial families.

Table 2- Summary statistics of the GNB dataset.

| GNB (Gram Negative Bacilli Bacteria) | | | | |
|---|---|---|---|---|
| Feature | Type | Feature | Type | |
| **Age (Years)** | Mean: 44.13<br>Std: 14.94<br>Range: (11-89) | **Organism** | ▪ *Citrobacter* (5.61%)<br>▪ *Acinetobacter spp* (7.48%)<br>▪ *Enterobacter spp*(1.87%)<br>▪ *Escherichia coli* (6.54%) | ▪ *Klebsiella spp* (28.04%)<br>▪ *Proteus spp* (4.67%)<br>▪ *Pseudomonas SPP* (9.34%)<br>▪ *Pseudomonas aeruginosa* (34.58%)<br>▪ *Stenotrophomonas maltophilia* (1.87%) |
| **Sex** | Male (54%)<br>Female (46%) | **Diagnosis** | ▪ Erythema Nodosum (0.93%)<br>▪ Hansen`s disease (35.51%)<br>▪ Mycetoma (0.93%) | ▪ Toxic Epidermal Necrolysis (2.80%)<br>▪ Abscess (0.93%)<br>▪ Burn (1.87%)<br>▪ Carbuncle (0.93%)<br>▪ Cellulitis (0.93%)<br>▪ Diabetic ulcer |
| **ESBL test** | ▪ Positive (26.17%)<br>▪ Negative (73.83%) | | | |
| **Carbapenems** | ▪ Positive (10.28%)<br>▪ Negative (89.71%) | | ▪ Pemphigus Vulgaris (5.61%)<br>▪ Gangrene (5.61%)<br>▪ Perianal ulcers (0.93%)<br>▪ Scrofuloderma (6.54%)<br>▪ Lupus Erythematosus (1.87%)<br>▪ Vascular ulcer (2.80%) | (1.87%)<br>▪ Ecthyma (1.87%)<br>▪ Eczema (2.80%)<br>▪ Furuncle (0.93%)<br>▪ Stasis ulcer (11.21%)<br>▪ Systemic sclerosis (0.93%)<br>▪ Traumatic ulcer (3.74%) |

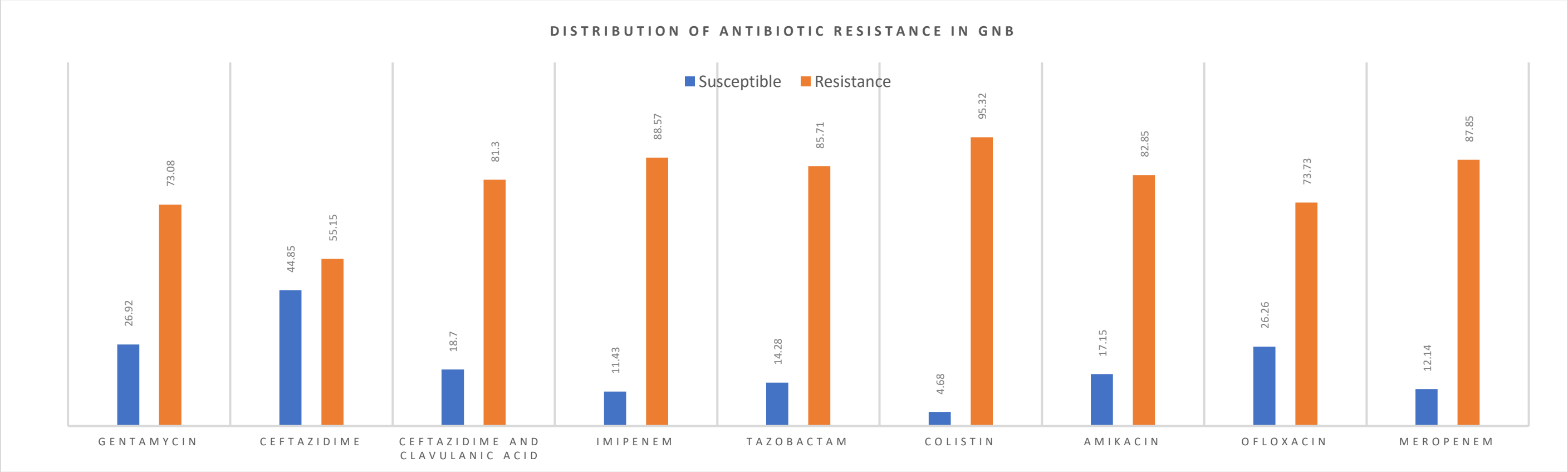

Figure 2- Percentage of GNB bacteria susceptible or resistant to nine antimicrobial families.

**Pearson Correlation Coefficient**

Pearson measures the direction of the linear relationship between two continuous variables with a normal distribution. The Pearson coefficients are between -1 and +1, in which positive signs indicate that the variables are linearly related by an increasing relationship with antibiotic resistance. In contrast, negative signs indicate that the variables are linearly related by a decreasing relationship with antibiotic resistance (increasing associated with antibiotic susceptibility). A strong relationship is considered with a coefficient greater than 0.8, while a weak correlation is in coefficients less than 0.5. The Pearson coefficients, $r$, can be calculated as follow:

$$r = \frac{n\sum xy - \sum x . \sum y}{\sqrt{[n\sum x^2 - (\sum x)^2] . [n\sum y^2 - (\sum y)^2}}$$

Where $x$ is the input variable, $y$ is the target variable indicating the antibiotic's resistance, and $n$ is the number of samples [12, 13].

**Spearman's correlation coefficient**

Spearman's correlation coefficient is a nonparametric statistical measure of the strength of a monotonic relationship between paired data. Unlike the Pearson correlation, the Spearman correlation does not assume that both datasets are normally distributed. Like the Pearson correlation, positive correlations imply that antibiotic resistance increases as x increases. Negative correlations indicate that as x increases, antibiotic susceptibility increases. The Spearman's Coefficients can be calculated from the following formula:

$$r_s = 1 - \frac{6\sum d_i^2}{n(n^2-1)}$$

Where $n$ is the number of samples, and $d_i$ is difference in ranks of the $i^{th}$ element [14, 15].

**Cramer's V analysis**

Cramer's V is based on the Pearson chi-square test and measures how strongly two categorical variables are associated. Unlike the Pearson and the Spearman correlation, The Cramer's V coefficients are between 0 and, +1 in which 0 means no association and 1 means full association. Cramer's coefficients, $\phi_c$, can be calculated from the following formula:

$$\phi_c = \sqrt{\frac{\chi^2}{N(k-1)}}$$

Where $\chi^2$ is the Pearson chi-square statistic, $N$ is the sample size, and $k$ is the lesser number of categories of either variable. Chi-square statistic, $\chi^2$, can be calculated as follow:

$$\chi^2 = \sum \frac{(X_o - X_e)^2}{X_e}$$

Where, $X_o$ is observed frequency and $X_e$ is expected frequency [16, 17].

## Antibiotic-Resistant classifiers

In this study, we propose machine learning (ML) techniques to classify antibiotic-resistant patients diagnosed with primary and secondary pyoderma based on their available data in the hospital information system, such as patient's demographics (age, gender), bacteria, and type of dermatoses (diagnosis). The ground truth labels were obtained from their previous antibiotic non-susceptibility testing.

We used the K-fold cross-validation technique with ten folds to train each classifier. Cross-validation is a reliable way to assess performance and reduce the variance of machine learning techniques. As noted in Table 1 and Table 2, we also performed bootstrapping techniques to over-sample the rare classes based on their distribution function to deal with highly imbalanced data. The cross-validation was done before oversampling to avoid overfitting [18, 19]. Figure 3 outlines data processing and classifiers' training protocol.

To assess classifiers performance, we considered the following metrics [20]:

a. **Sensitivity or Recall:** Determines the proportion of instances classified as resistant to antibiotics correctly.
b. **Precision:** Determines the proportion of relevant instances, which measures the quality of the positive predictions made by the model.
c. **F-2-score**: combines the precision and sensitivity of the model. It is possible to adjust the F-score to give more weights to precision over recall (F-0.5-score) or vice versa (F-2-score). F-1-score gives equal weight to both. In antibiotic-resistant studies, false negatives are a more serious concern than false positives. False negatives indicate the samples are resistant to antibiotics but classified as susceptible. This misclassification can put a patients' life in danger; therefore, we calculated the F-2 score for each classifier to put more attention on minimizing false negatives.
d. **Area Under the Receiver Operating Characteristics Curve (AUC-ROC):** Measures the performance of a binary classifier to distinguish between classes at various thresholds. ROC is a probability curve that plots the True Positive Rates (TPR) against False Positive Rates (FPR) at different thresholds, and AUC shows the degree of separability. TPR is the proportion of resistant instances that the model correctly classifies, whereas FPR is of susceptible samples misclassified as resistant. AUC= 1 represents a perfect classifier, AUC=0.5 is a random choice.

We trained three different classifiers: Random Forest, Multilayer Perceptron Neural Network, and Convolutional Neural Network for each antibiotic target in GPC and GNB dataset.

**Random Forest (RF)**

Random forest developed by Leo Breiman is an ensemble learner that generates multiple classifiers and aggregates their results. RF classifier is a set of decision trees created from a randomly selected training set. Each tree in RF will vote for its input, and the majority voting of trees determines the output. In RFs, each tree makes a class prediction, and the class with the most votes becomes the final model's prediction [21]. Each tree in RFs is grown using a subset of training samples, and some variables which are not used to grow the corresponding trees are known as out-of-bag (OOB) samples. One of the properties of OOBs is the estimation of variable importance, which quantifies the degree of contribution of a given variable in providing classification accuracy [22]. This is one of the advantages of RF over deep models, which is done by measuring the misclassification rate when the OOB examples for a variable, $x_i$, are randomly permuted and passed through the corresponding tree to vote for $x_i$. If the classification accuracy decreases significantly, it suggests a substantial contribution of variable $x_i$ in the classification result. On the other hand, if it doesn't affect the predictive performance, then $x_i$ is considered unimportant [23].

**Multilayer Perceptron (MLP)**

MLP is an artificial neural network model that maps input data to a set of suitable outputs through multiple nonlinear or linear functions. MLP is supervised learning in which, during training, the weights are updated by the backpropagation algorithm and by propagating the errors backward from the output layer to the input layer. The performance of MLP is highly dependent on hyperparameters: number of neurons, number of hidden layers, learning rate, and momentum [24]. MLP has been widely used in medical applications such as heart disease diagnosis [25], thyroid disease diagnosis [26], breast cancer classification [27], and chronic kidney disease prediction [28]. In this study, we trained a two-layer feedforward neural network with a structure of (16,1). The activation function of the first layer is ReLU, and the output layer is sigmoid.

**Convolutional Neural Network (CNN)**

ConvNets or CNNs can process data that come in multiple arrays, such as speech, text, image, and video. ConvNets are constructed in various stages. The first stage is the convolutional layer, in which a set of weights called a filter bank is convolved with the input vector. This locally weighted sum is then passed through a nonlinearity such as a ReLU called activation function. Two or three stages of convolution, activation functions, and pooling layers are stacked, followed by fully connected layers. Back propagating gradients through a ConvNet train all the filter banks' weights. This hierarchy structure will allow the higher-level features (details) to be obtained by composing a lower-level one. The convolutional and pooling layers in ConvNets are inspired by the cells in visual neuroscience [29]. There have been numerous applications of convolutional networks in medicine, such as medical image segmentation[30, 31], wound image classification [32, 33], breast cancer classification and diagnosis[34], and medical image denoising and enhancement [35, 36]. Our CNN classifier comprises one 1-D Convolution layer followed by four dense layers. The convolution layer has 64 filters of size three. The activation functions are ReLU except for the last layer, which is a sigmoid function.

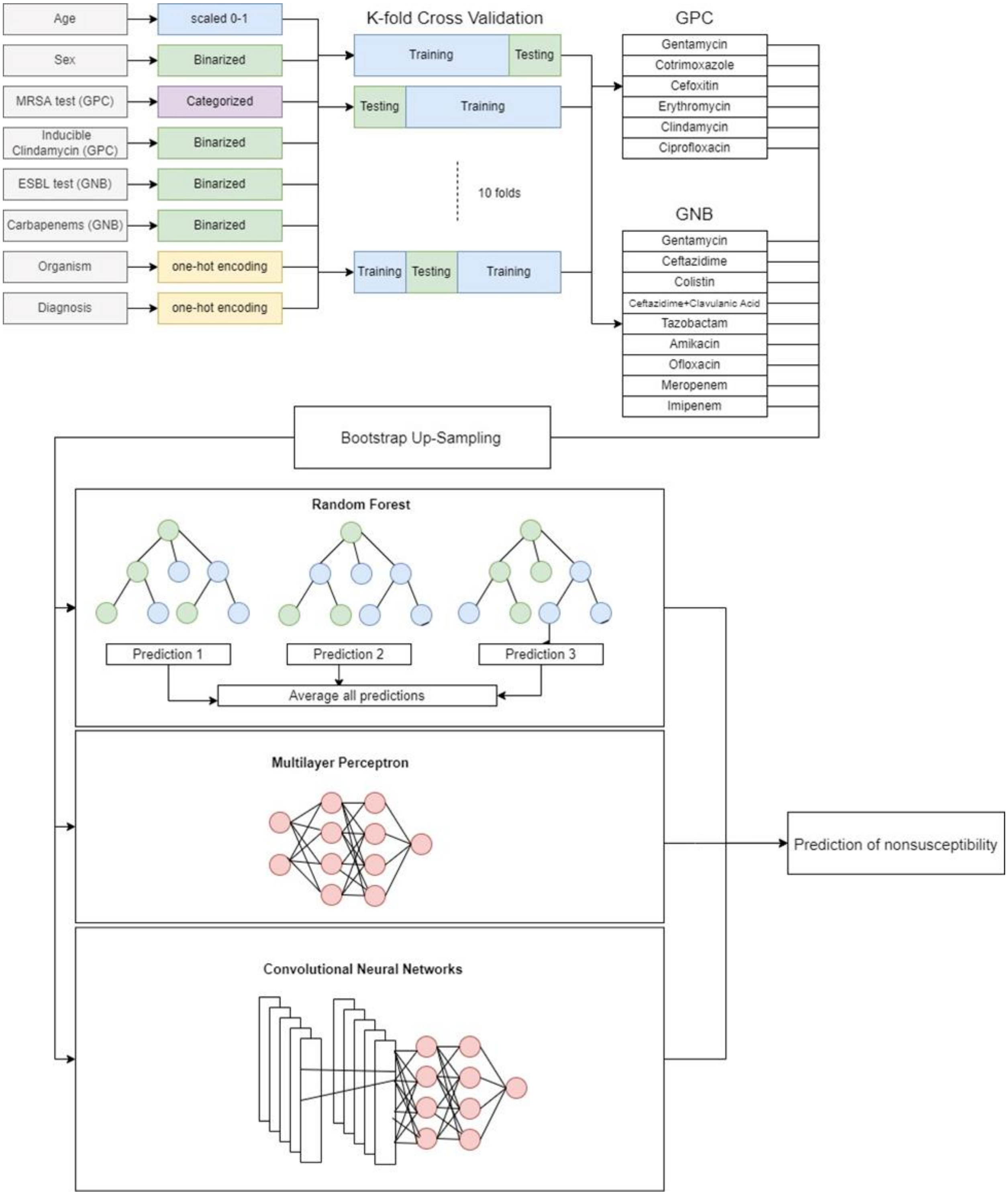

Figure 3- Data Processing includes scaling, categorization, and one-hot encoding. K-fold cross-validation is applied ten times by randomly selecting training and testing set by the proportion of 0.8 to 0.2, respectively. Dataset was labeled resistant or susceptible to each antibiotic family (six families for GPC and nine for GNB). Due to having highly imbalanced data, bootstrap up-sampling is applied to up sample the smaller class. A Random Forest, Multilayer Perceptron, and Convolution Neural Network is trained for each antibiotic family on each k-fold dataset. The networks were tested on the test set, and the metrics indicate the average of 10-folds.

## Results

***Gram-Positive Cocci (GPC)***: we used correlation and statistical analysis to find the association between organism, diagnosis, age, and sex with the families of antibiotics. Figure 4 represents the correlation coefficients of each factor with the six antimicrobial families. Stars represent variables with significant correlation to antibiotic resistance (P-values < 0.05). According to Figure 4, the MRSA screening test revealed the highest correlation in predicting the outcome of all six antibiotics, followed by the inducible clindamycin resistance test, which showed a significant association in predicting Gentamicin, Cefoxitin, Erythromycin, and Clindamycin resistance. Enterococcus SPP was found to be highly resistant to Cotrimoxazole. Staphylococcus aureus, coagulase-negative Staphylococcus, and Streptococcus pyogenes were significantly associated with Cotrimoxazole and Ciprofloxacin resistance. Surprisingly, diagnoses showed no association with antibiotic resistance patterns. Increasing age and female gender were found to be highly correlated with Ciprofloxacin resistance.

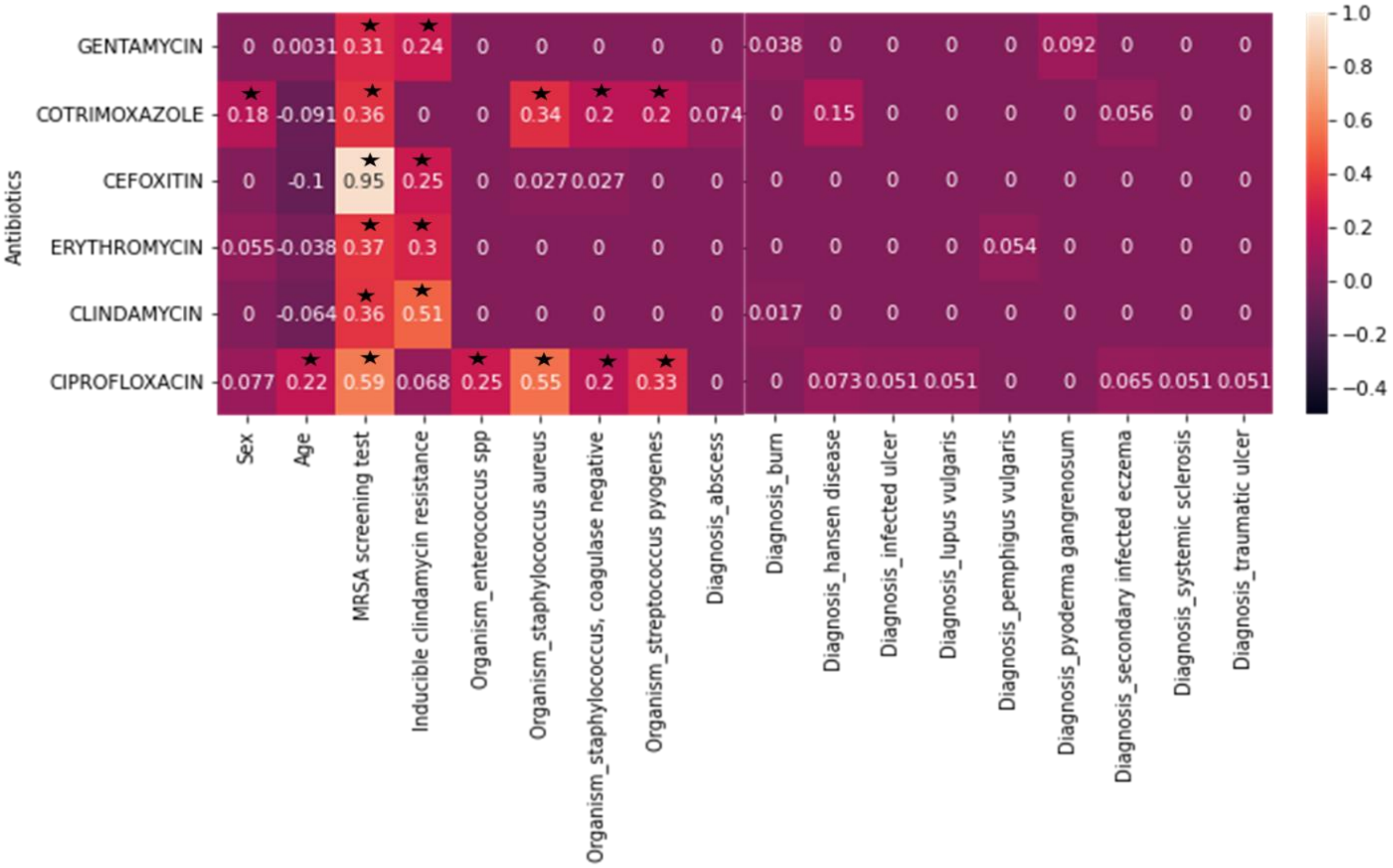

Figure 4- Correlation coefficients between each factor and six antibiotic families in samples with GPC bacteria. Stars represent a significant linear correlation.

Table 3 shows the performance results for classifying Gentamicin, Cotrimoxazole, Cefoxitin, Erythromycin, Clindamycin, and Ciprofloxacin resistance obtained from Random Forest (RF), Multilayer Perceptron (MLP), and Convolutional Neural Network (CNN) by using 10-fold cross-validation. Generally, Cefoxitin and Ciprofloxacin resistance was more predictable than the other antibiotics families by F-2score and an AUC of more than 0.9. Comparing different algorithms, MLP revealed a higher AUC and F-2 scores in predicting Gentamicin, Cotrimoxazole resistance, and relatively comparable performance in Cefoxitin, Clindamycin, and Ciprofloxacin resistance as compared to CNN.

Table 3- Performance (Recall, Precision, F-2 score, AUC) provided by three models (RF, MLP, CNN) on the test dataset. The purpose is to determine the resistance of GPC to six families of antimicrobials. Bolds indicate the highest AUC in each family.

| Family | Model | Recall | Precision | F-2 Score | AUC |
|---|---|---|---|---|---|
| **Gentamicin** | **RF** | 0.62 | 0.54 | 0.60 | 0.65 |
| | **MLP** | 0.64 | 0.52 | 0.61 | **0.68** |
| | **CNN** | 0.60 | 0.48 | 0.57 | 0.60 |
| **Cotrimoxazole** | **RF** | 0.67 | 0.55 | 0.64 | **0.71** |
| | **MLP** | 0.68 | 0.56 | 0.65 | **0.71** |
| | **CNN** | 0.55 | 0.42 | 0.52 | 0.70 |
| **Cefoxitin** | **RF** | 0.99 | 0.92 | 0.97 | 0.96 |
| | **MLP** | 0.99 | 0.92 | 0.97 | **0.98** |
| | **CNN** | 0.98 | 0.93 | 0.97 | **0.98** |
| **Erythromycin** | **RF** | 0.81 | 0.78 | 0.80 | 0.59 |
| | **MLP** | 0.82 | 0.83 | 0.82 | 0.70 |
| | **CNN** | 0.78 | 0.86 | 0.80 | **0.73** |
| **Clindamycin** | **RF** | 0.80 | 0.66 | 0.77 | 0.68 |
| | **MLP** | 0.76 | 0.69 | 0.74 | **0.76** |
| | **CNN** | 0.76 | 0.70 | 0.75 | **0.76** |
| **Ciprofloxacin** | **RF** | 0.94 | 0.95 | 0.94 | 0.82 |
| | **MLP** | 0.92 | 0.96 | 0.93 | 0.90 |
| | **CNN** | 0.91 | 0.95 | 0.92 | **0.91** |

The first ten important features calculated by the Random Forest classifier is illustrated in Figure 5. Although MLP and CNN generally outperformed RF in predicting antibiotic resistance, Random Forest had the advantage of calculating the importance of features contributing to the outcome, as explained in the previous section. Age was among the top two critical features to predict antibiotic resistance. MRSA screening test contributed to Cefoxitin resistant prediction by a score of 0.6. As expected, Inducible Clindamycin resistance was the most critical predictor in Clindamycin resistance prediction.

***Gram-Negative Bacilli (GNB)***: Figure 6 represents the correlation coefficients of each factor (organism, diagnosis, age, sex, ESBL, and carbapenems) with the nine antimicrobial families (Gentamicin, Amikacin, Ceftazidime, Ceftazidime + Clavulanic Acid, Imipenem, Piperacillin + Tazobactam, Colistin, Ofloxacin, and Meropenem). Stars represent variables with significant correlation to antibiotic resistance (P-values $< 0.05$). Age was found to be significantly correlated with Ceftazidime and Piperacillin + Tazobactam resistance. More interestingly, unlike GPC, diagnoses highly contributed to the prediction of antimicrobial resistance in GNB. Proteus organisms with resistance to Colistin showed the highest correlation coefficients among all the factors (0.9 and 0.89, respectively).

Table 4 provides the classification performance of RF, MLP, and CNN in nine antibiotic families. In general, the performance depended on the family of antimicrobials. However, CNN outperformed the other classification techniques in most of the families. Results suggest that nonlinear relationships are essential to identify antimicrobial resistance in GNB samples. Gentamicin and Amikacin have shown the lowest prediction accuracy by machine learning, with the F-2 score worse than random selection. On the other hand, Colistin, Imipenem, Ceftazidime, Ofloxacin, and Meropenem have shown high performance (AUC > 0.87, F-2 score > 0.77 and Precision > 0.83).

Figure 7 represents the first ten important features analyzed by the RF classifier in GNB samples. As GPC, age was the first or the second important feature to predict antibiotic resistance in GNB samples except for Colistin.

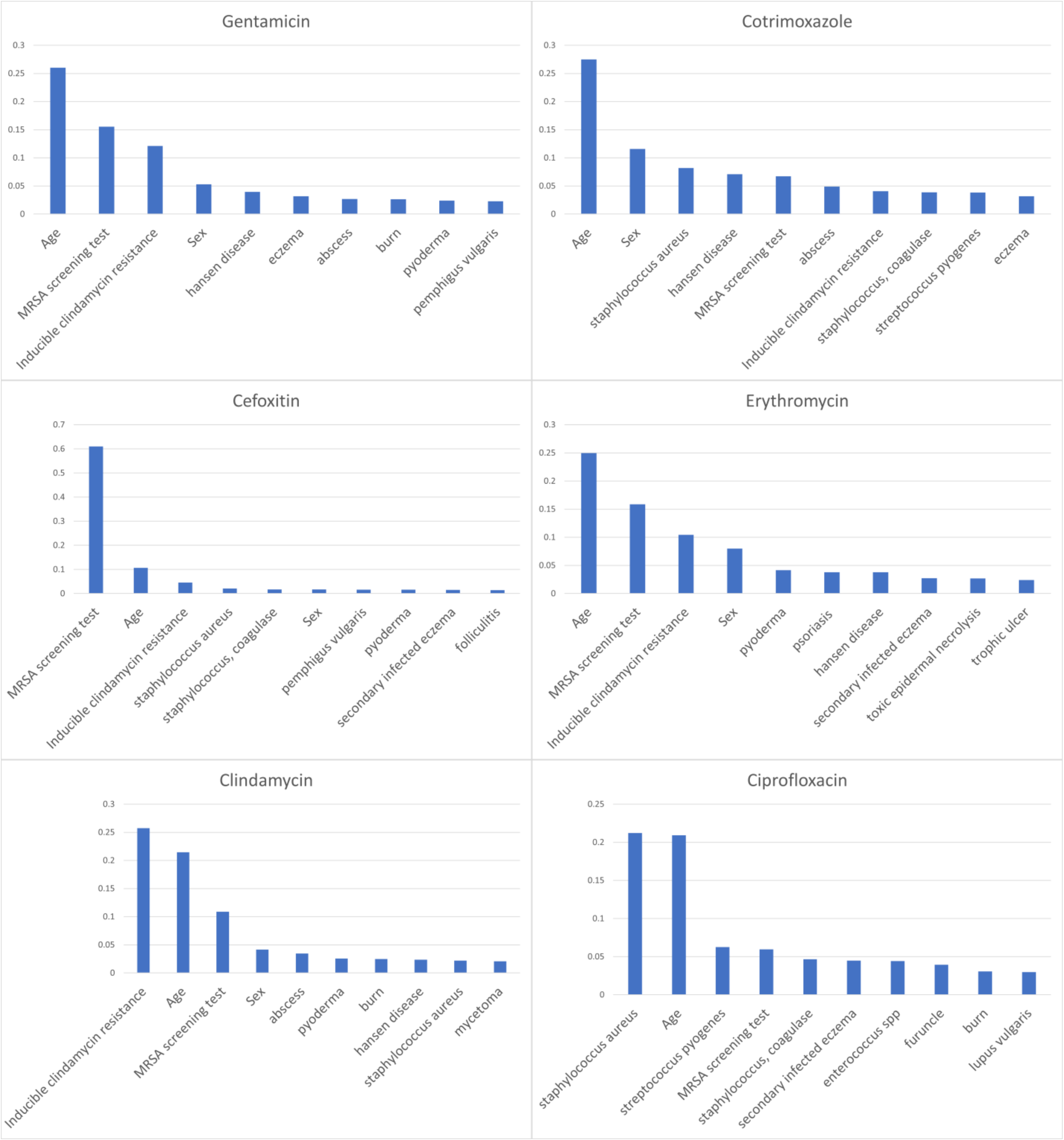

Figure 5-Relative Importance of Feature (RIF) analysis by RF classifier in GPC samples.

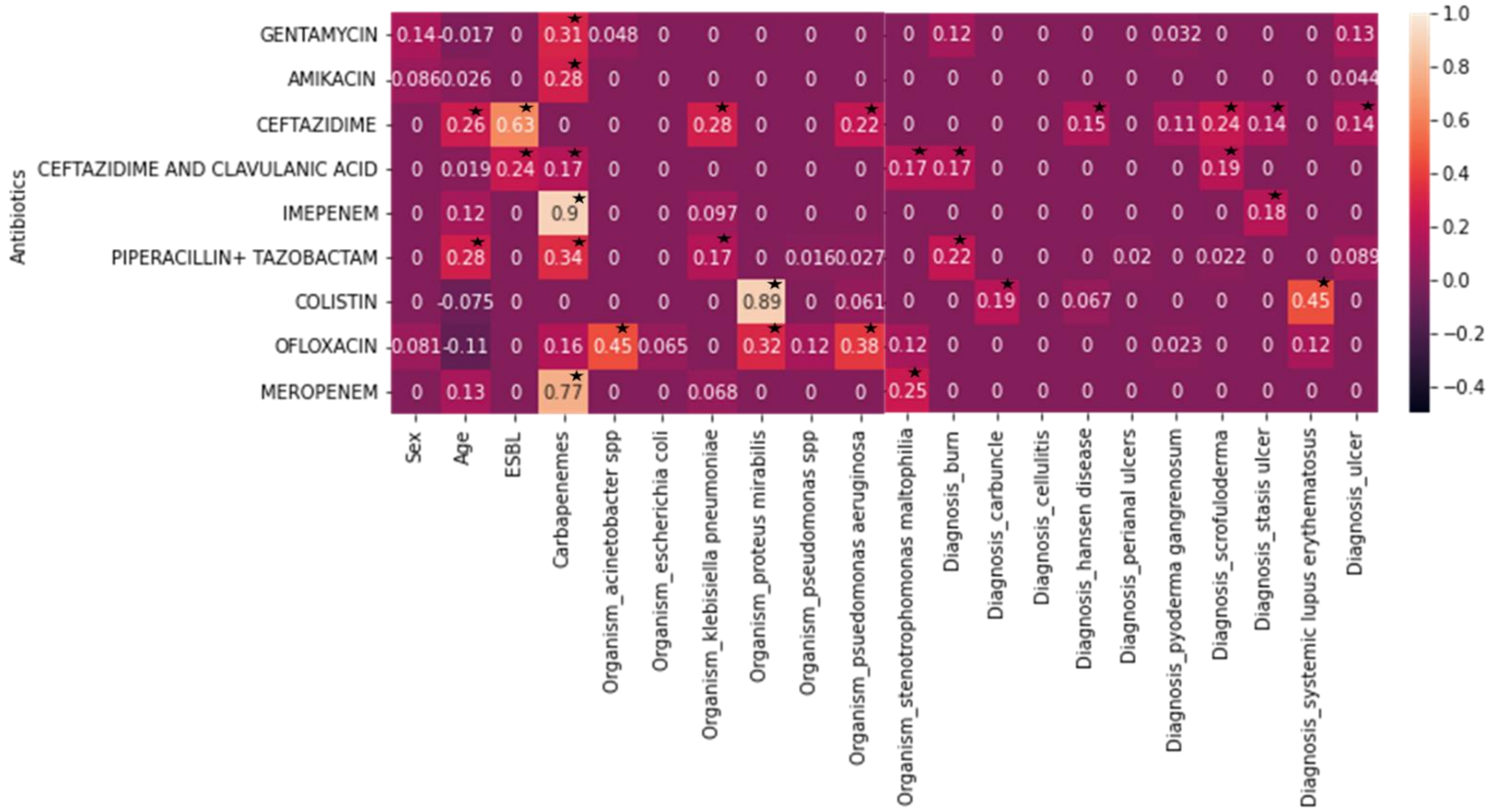

Figure 6- Correlation coefficients between each factor and nine antibiotic families in samples with GNB bacteria. Stars represent a significant linear correlation.

Table 4- Performance (Recall, Precision, F-2 score, AUC) provided by three models (RF, MLP, CNN) on the test dataset. The purpose is to determine the resistance of GNB to nine families of antimicrobials. Bolds indicate the highest AUC in each family.

| Family | Model | Recall | Precision | F-2 Score | AUC |
|---|---|---|---|---|---|
| **Gentamicin** | **RF** | 0.43 | 0.48 | 0.44 | **0.62** |
| | **MLP** | 0.44 | 0.37 | 0.42 | **0.62** |
| | **CNN** | 0.54 | 0.35 | 0.48 | 0.61 |
| **Amikacin** | **RF** | 0.25 | 0.19 | 0.23 | 0.55 |
| | **MLP** | 0.48 | 0.22 | 0.33 | 0.51 |
| | **CNN** | 0.37 | 0.35 | 0.36 | **0.56** |
| **Ceftazidime** | **RF** | 0.80 | 0.88 | 0.81 | 0.85 |
| | **MLP** | 0.84 | 0.91 | 0.85 | 0.86 |
| | **CNN** | 0.82 | 0.83 | 0.82 | **0.87** |
| **Ceftazidime + Clavulanic Acid** | **RF** | 0.42 | 0.54 | 0.44 | 0.66 |
| | **MLP** | 0.55 | 0.45 | 0.50 | **0.76** |
| | **CNN** | 0.53 | 0.43 | 0.49 | 0.73 |
| **Imipenem** | **RF** | 0.83 | 1 | 0.86 | 0.92 |
| | **MLP** | 0.87 | 1 | 0.88 | 0.93 |
| | **CNN** | 0.87 | 1 | 0.88 | **0.93** |
| **Piperacillin + Tazobactam** | **RF** | 0.62 | 0.56 | 0.61 | 0.78 |
| | **MLP** | 0.56 | 0.45 | 0.51 | 0.83 |
| | **CNN** | 0.78 | 0.53 | 0.68 | **0.86** |
| **Colistin** | **RF** | 0.9 | 1 | 0.92 | 0.95 |
| | **MLP** | 1 | 1 | 1 | **0.99** |
| | **CNN** | 1 | 1 | 1 | **0.99** |
| **Ofloxacin** | **RF** | 0.68 | 0.82 | 0.70 | 0.81 |
| | **MLP** | 0.77 | 0.83 | 0.77 | **0.89** |
| | **CNN** | 0.65 | 0.81 | 0.66 | 0.85 |
| **Meropenem** | **RF** | 0.79 | 0.93 | 0.81 | 0.89 |
| | **MLP** | 0.90 | 0.91 | 0.89 | 0.90 |
| | **CNN** | 0.88 | 0.93 | 0.88 | 0.94 |

## Discussion

SSTIs are treated based on risk factors, clinical presentation, disease severity, hospital environment, prior antibiotic therapy with culture and antibiotic susceptibility testing. In the present study, we have demonstrated that machine learning algorithms can determine if an SSTI is resistant to an antibiotic by using a dataset that contains demographic and clinical data such as age, sex, diagnosis, and the bacterial pathogens involved with skin infections. Our results showed that ML techniques might help dermatologists with targeted empiric antibiotic choices. We compared the accuracy using AUC, Recall, Precision, and F-2 scores in deep and conventional ML algorithms and showed that the MLP and CNN outperformed RF in predicting antibiotic resistance in most antibiotic families. Generally, the deep learning models have shown better prediction performance than the conventional algorithms by capturing the nonlinear relationship between the datasets. However, conventional algorithms helped determine the relative importance of each predictor in the outcome. Despite the poor performance of RF in nonsusceptibility prediction, we employed it to study the contribution of each factor in prediction and some conventional correlation analysis. However, the observations can be highly related to the dataset size and the class distribution [37].

It is reported that any model with AUC> 0.7 can be regarded as a good fit [38]. We achieved the AUC> 0.7 in all the families except Gentamicin in GPC and Gentamicin and Amikacin in GNB samples. Generally, it is more challenging to predict resistance for antibiotics associated with unknown or multifactorial resistance mechanisms than those in which resistance is significantly related to a single variable such as Colistin[37]. We achieved an AUC of 0.99 using MLP and CNN to predict Colistin resistance due to the presence of Proteus mirabilis, a bacteria intrinsically resistant to Colistin[39]. Our correlation analysis also confirmed the same by revealing a significant correlation between Proteus mirabilis and Colistin resistance with a coefficient of 0.89. Proteus mirabilis was also detected as the most crucial feature in predicting Colistin resistance with the coefficient of 0.47 due to its intrinsic characteristics. It is also reported that Enterococcus is intrinsically resistant to the Cotrimoxazole and Clindamycin[40]; however, neither the correlation analysis nor the relative importance of features using RF showed any significant correlation between Enterococcus and the Cotrimoxazole and Clindamycin resistance. The classification results indicated the maximum AUC of 0.71 and 0.78 in the prediction of Cotrimoxazole and Clindamycin resistance. A possible reason is the low distribution of Enterococcus in our dataset (2/103 = 1.94. Age was also shown to be a decisive factor in predicting resistance to most antibiotic families. Garcia et al. investigated the correlation between age and antibiotic resistance in patients with positive MRSA and found that the antibiotics that target DNA syntheses, such as Ofloxacin and Ciprofloxacin, show a significant correlation between older patients and antibiotic resistance. However, antibiotics that target ribosomal functions or cell wall synthesis, such as aminoglycosides, cephalosporins etc., showed a consistent degree of resistance across all age classes[41].

In the present study, the most common GNB were non-fermenters like *Pseudomonas aeruginosa and Acinetobacter spp*. (53.71%) compared to Enterobacteriaceae (46.72%).Ramkrishna et al., though, had found more Enterobacteriaceae than non-fermenters [42]. Pseudomonas aeruginosa (34.58%) was the most common gram-negative organism susceptible to Piperacillin+Tazobactum (91.89). In their study, Siami et al. had also reported the effectiveness of piperacillin+Tazobactum in SSTI [43]. Acinetobacter baumannii (7.48%) complex, susceptible to carbapenems and cephalosporins, was found in significantly less numbers in our study. At the same time, Guerrero et al. reported MDR while Jabbour J et a. found XDR Acinetobacter baumanni from SSTI in their study. Among the Gram-negative bacilli isolated, the overall resistance rates were found to be the highest with Klebsiella spp (28.04%) followed by Acinetobacter spp but reverse results were observed with studies from various parts of India [44]. Staphylococcus aureus was the most common Gram-positive isolate (82.52%) of which around 38% were found to be MRSA. On the other hand, only 10% of the isolates were Coagulase negative Staphylococcus species. There findings were similar to previous studies done by Drydan et al. [45] and Stryjewski et al. [46].

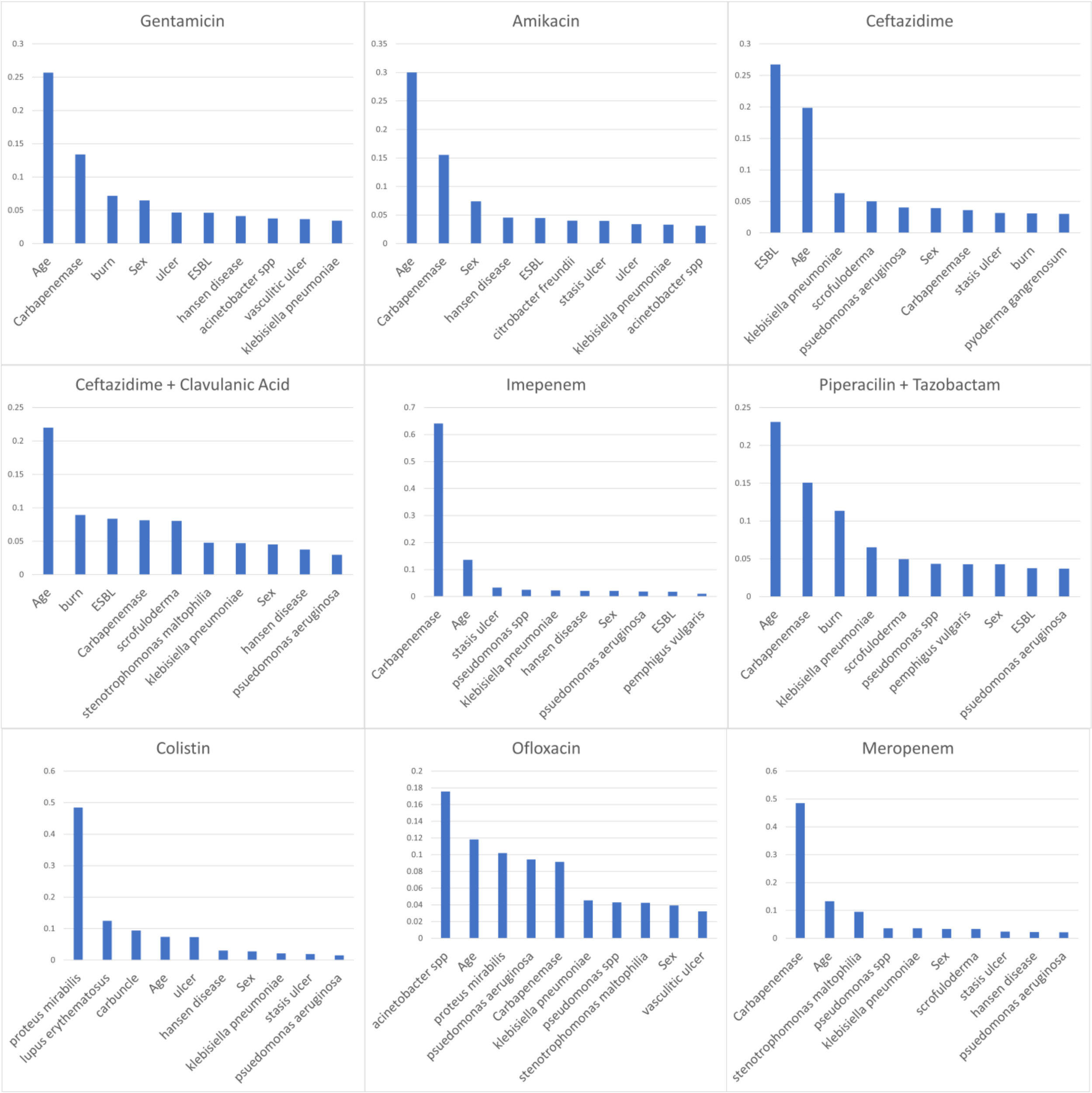

Figure 7 -Relative Importance of Feature (RIF) analysis by RF classifier in GNB samples.

While ML techniques have shown high accuracy in prediction of antibiotics' nonsusceptibility, a more comprehensive assessment is needed to build highly accurate and reliable models. The performance of the ML algorithms is largely dependent on the dataset size and metrics. Since the number of data points was limited and highly imbalanced in the present study, the performance of the classification techniques can be attributed to the class distribution. Our future work will employ a different strategy to balance classes based on generative adversarial networks instead of an up-sampling bootstrapping strategy. Since data collection is costly and time-consuming, this approach could generate an augmented dataset to increase the number of training samples and improve the classification performance [47].

## References

1. Mims, C., et al., *Medical microbiology.* Structure, 2004. **7**: p. 7.
2. Oliveira, J. and W.C. Reygaert, *Gram negative bacteria.* 2019.
3. Vazquez-Guillamet, C. and M.H. Kollef, *Treatment of gram-positive infections in critically ill patients.* BMC infectious diseases, 2014. **14**(1): p. 1-8.
4. Dryden, M., et al., *A European survey of antibiotic management of methicillin-resistant Staphylococcus aureus infection: current clinical opinion and practice.* Clinical Microbiology and Infection, 2010. **16**: p. 3-30.
5. Stevens, D.L., et al., *Practice guidelines for the diagnosis and management of skin and soft-tissue infections.* Clinical Infectious Diseases, 2005. **41**(10): p. 1373-1406.
6. DiNubile, M.J. and B.A. Lipsky, *Complicated infections of skin and skin structures: when the infection is more than skin deep.* Journal of Antimicrobial Chemotherapy, 2004. **53**(suppl_2): p. ii37-ii50.
7. Jacobs, E., A. Dalhoff, and G. Korfmann, *Susceptibility patterns of bacterial isolates from hospitalised patients with respiratory tract infections (MOXIAKTIV Study).* International journal of antimicrobial agents, 2009. **33**(1): p. 52-57.
8. May, A.K., et al., *Treatment of complicated skin and soft tissue infections.* Surgical infections, 2009. **10**(5): p. 467-499.
9. Jones, M.E., et al., *Epidemiology and antibiotic susceptibility of bacteria causing skin and soft tissue infections in the USA and Europe: a guide to appropriate antimicrobial therapy.* International journal of antimicrobial agents, 2003. **22**(4): p. 406-419.
10. Kumar, S. and I. Chong, *Correlation analysis to identify the effective data in machine learning: Prediction of depressive disorder and emotion states.* International journal of environmental research and public health, 2018. **15**(12): p. 2907.
11. Peykani, P., et al., *A novel mathematical approach for fuzzy multi-period multi-objective portfolio optimization problem under uncertain environment and practical constraints.* Journal of fuzzy extension and application, 2021. **2**(3): p. 191-203.
12. Benesty, J., et al., *Pearson correlation coefficient*, in *Noise reduction in speech processing*. 2009, Springer. p. 1-4.
13. Sedgwick, P., *Pearson's correlation coefficient.* Bmj, 2012. **345**.
14. Zar, J.H., *Spearman rank correlation.* Encyclopedia of biostatistics, 2005. **7**.
15. Zar, J.H., *Significance testing of the Spearman rank correlation coefficient.* Journal of the American Statistical Association, 1972. **67**(339): p. 578-580.
16. McHugh, M.L., *The chi-square test of independence.* Biochemia medica, 2013. **23**(2): p. 143-149.
17. Khamechian, M. and M.E. Petering, *A Mathematical Modeling Approach to University Course Planning.* Computers & Industrial Engineering, 2021: p. 107855.
18. Yap, B.W., et al. *An application of oversampling, undersampling, bagging and boosting in handling imbalanced datasets*. in *Proceedings of the first international conference on advanced data and information engineering (DaEng-2013)*. 2014. Springer.
19. Thanathamathee, P. and C. Lursinsap, *Handling imbalanced data sets with synthetic boundary data generation using bootstrap re-sampling and AdaBoost techniques.* Pattern Recognition Letters, 2013. **34**(12): p. 1339-1347.
20. Kasperczuk, A. and A. Dardzińska, *Comparative evaluation of the different data mining techniques used for the medical database.* acta mechanica et automatica, 2016. **10**(3).
21. Breiman, L., *Random forests.* Machine learning, 2001. **45**(1): p. 5-32.
22. Khalilia, M., S. Chakraborty, and M. Popescu, *Predicting disease risks from highly imbalanced data using random forest.* BMC medical informatics and decision making, 2011. **11**(1): p. 1-13.

23. Sage, A., *Random forest robustness, variable importance, and tree aggregation.* 2018.
24. Taud, H. and J. Mas, *Multilayer perceptron (MLP)*, in *Geomatic Approaches for Modeling Land Change Scenarios*. 2018, Springer. p. 451-455.
25. Yan, H., et al., *A multilayer perceptron-based medical decision support system for heart disease diagnosis.* Expert Systems with Applications, 2006. **30**(2): p. 272-281.
26. Hosseinzadeh, M., et al., *A multiple multilayer perceptron neural network with an adaptive learning algorithm for thyroid disease diagnosis in the internet of medical things.* The Journal of Supercomputing, 2021. **77**(4): p. 3616-3637.
27. Ting, F. and K. Sim. *Self-regulated multilayer perceptron neural network for breast cancer classification*. in *2017 International Conference on Robotics, Automation and Sciences (ICORAS)*. 2017. IEEE.
28. Yildirim, P. *Chronic kidney disease prediction on imbalanced data by multilayer perceptron: Chronic kidney disease prediction*. in *2017 IEEE 41st Annual Computer Software and Applications Conference (COMPSAC)*. 2017. IEEE.
29. LeCun, Y., Y. Bengio, and G. Hinton, *Deep learning.* nature, 2015. **521**(7553): p. 436-444.
30. Gheshlaghi, S.H., et al. *Efficient Oct Image Segmentation Using Neural Architecture Search*. in *2020 IEEE International Conference on Image Processing (ICIP)*. 2020. IEEE.
31. Dehzangi, O., et al. *OCT Image Segmentation Using Neural Architecture Search and SRGAN*. in *2020 25th International Conference on Pattern Recognition (ICPR)*. 2021. IEEE.
32. Rostami, B., et al., *Multiclass Wound Image Classification using an Ensemble Deep CNN-based Classifier.* Computers in Biology and Medicine, 2021: p. 104536.
33. Anisuzzaman, D., et al., *Multi-modal Wound Classification using Wound Image and Location by Deep Neural Network.* arXiv preprint arXiv:2109.06969, 2021.
34. Gheshlaghi, S.H., C.N.E. Kan, and D.H. Ye. *Breast Cancer Histopathological Image Classification with Adversarial Image Synthesis*. in *2021 43rd Annual International Conference of the IEEE Engineering in Medicine & Biology Society (EMBC)*. 2021. IEEE.
35. Jifara, W., et al., *Medical image denoising using convolutional neural network: a residual learning approach.* The Journal of Supercomputing, 2019. **75**(2): p. 704-718.
36. Zhang, Y. and H. Yu, *Convolutional neural network based metal artifact reduction in x-ray computed tomography.* IEEE transactions on medical imaging, 2018. **37**(6): p. 1370-1381.
37. Hicks, A.L., et al., *Evaluation of parameters affecting performance and reliability of machine learning-based antibiotic susceptibility testing from whole genome sequencing data.* PLoS computational biology, 2019. **15**(9): p. e1007349.
38. Cho, S.K., et al., *Development of a model to predict healing of chronic wounds within 12 weeks.* Advances in wound care, 2020. **9**(9): p. 516-524.
39. Samonis, G., et al., *Trends of isolation of intrinsically resistant to colistin Enterobacteriaceae and association with colistin use in a tertiary hospital.* European journal of clinical microbiology & infectious diseases, 2014. **33**(9): p. 1505-1510.
40. Olawale, K.O., S.O. Fadiora, and S.S. Taiwo, *Prevalence of hospital acquired enterococci infections in two primary-care hospitals in Osogbo, Southwestern Nigeria.* African journal of infectious diseases, 2011. **5**(2).
41. Garcia, A., T. Delorme, and P. Nasr, *Patient age as a factor of antibiotic resistance in methicillin-resistant Staphylococcus aureus.* Journal of medical microbiology, 2017. **66**(12): p. 1782-1789.
42. Sivaramakrishna, D., et al., *Pretreatment with KOH and KOH-urea enhanced hydrolysis of α-chitin by an endo-chitinase from Enterobacter cloacae subsp. cloacae.* Carbohydrate polymers, 2020. **235**: p. 115952.
43. Siami, G., et al., *Clinafloxacin versus piperacillin-tazobactam in treatment of patients with severe skin and soft tissue infections.* Antimicrobial agents and chemotherapy, 2001. **45**(2): p. 525-531.

44. Manchanda, V., et al., *Multidrug resistant acinetobacter*. J Glob Infect Dis. 2010;**2**(3):291-304. doi:10.4103/0974-777X.68538

45. Drydan, M. S., et al., *Complicated skin and soft tissue infection*, Journal of Antimicrobial Chemotherapy, Volume 65, Issue suppl_3, November 2010, Pages iii35–iii44

46. Stryjewski, M.E., et al., *Skin and soft-tissue infections caused by community-acquired methicillin-resistant Staphylococcus aureus.* Clin Infect Dis. 2008 Jun 1;46 Suppl 5:S368-77. doi: 10.1086/533593. PMID: 18462092.

47. Foomani, F.H., et al., *Synthesizing time-series wound prognosis factors from electronic medical records using generative adversarial networks.* Journal of biomedical informatics, 2022. **125**: p. 103972.